\documentclass[letterpaper]{article}

\usepackage{aaai}
\usepackage{times}
\usepackage{helvet}
\usepackage{courier}

\usepackage{amsmath}

\usepackage{graphicx}
\usepackage{booktabs}
\usepackage{enumerate}
\usepackage[table]{xcolor}

\DeclareMathOperator*{\argmax}{argmax}
\newcommand{\G}{\cellcolor[gray]{0.75}}

\usepackage{url}

\newcommand{\Omit}[1]{}

\begin{document}
% The file aaai.sty is the style file for AAAI Press 
% proceedings, working notes, and technical reports.
%
\title{Action Selection for MDPs: Anytime AO* vs.\ UCT}
% \author{\textbf{ID: \# 561}  \  \ \ \ \  \textbf{Keywords:} MDPs, Heuristic Search, UCT}

\author{Blai Bonet \\
%         Departamento de Computaci\'on y TI \\
        Universidad Sim\'on Bol\'{\i}var \\
        Caracas, Venezuela \\
        {\normalsize\url{bonet@ldc.usb.ve}}
\And
        Hector Geffner \\
%         Department of Information and Commnication Technologies (DTIC) \\
        ICREA \&  Universitat Pompeu Fabra \\
        08003  Barcelona, SPAIN \\
        {\normalsize\url{hector.geffner@upf.edu}}}

\maketitle
\begin{abstract}
\begin{quote}
In the presence of non-admissible heuristics, A* and other best-first algorithms 
can be  converted into anytime optimal algorithms over OR graphs, by simply continuing
the search after the first solution is found. The same trick, however, does not work 
for  best-first algorithms over AND/OR graphs, that must be able to expand leaf nodes of 
the explicit graph that are not necessarily part of the best partial solution.
Anytime optimal variants of AO* must thus address an exploration-exploitation tradeoff: 
they cannot just "exploit", they must keep exploring as well. In this work,  
we develop one such variant of AO* and apply it to finite-horizon MDPs.  
This Anytime AO* algorithm eventually delivers an optimal policy  while using
%HG
\emph{non-admissible random heuristics that can be sampled}, as when the heuristic
is the cost of a base policy that can be sampled with rollouts.
We then test Anytime AO* for action selection
over large infinite-horizon MDPs that cannot be solved with existing off-line
heuristic search and dynamic programming algorithms, and compare it with UCT.
\end{quote}
\end{abstract}

\section{Introduction}

One of the natural approaches for selecting  actions in  very large state spaces
is by performing a limited amount of lookahead. In the contexts of discounted
MDPs, Kearns, Mansour, and Ng have shown  that near to optimal actions can be selected by considering a 
\emph{sampled}  lookahead tree that is sufficiently \emph{sparse},   whose size depends on the discount factor
and the suboptimality bound  but not on  the number of problem states \cite{kearns:sparse99}. 
The UCT algorithm \cite{uct} is a version of this form of Monte Carlo planning, where the lookahead trees are not grown
depth-first but `best-first',  following a selection criterion that balances `exploration' and `exploitation'
borrowed from the UCB algorithm  for  multi-armed bandit problems \cite{ucb}.

While UCT does not inherit the theoretical properties of either Sparse Sampling or UCB,
UCT is an \emph{anytime optimal algorithm} for discounted or finite horizon MDPs that
eventually picks up the optimal actions when given sufficient time. The popularity of 
%HG
% UCT, however, is not due to this property but to its success in the game of Go,
UCT follows from its success in the game of Go where it outperformed all other approaches \cite{uct:go}; a success that
has been replicated in other models and tasks such as   Real-Time Strategy Games \cite{fern:uct}, 
General Game Playing \cite{uct:ggp}, and POMDPs \cite{silver:pomdps}.

An original  motivation for the work reported in this paper was to get a better understanding
of  the success of UCT and related Monte-Carlo Tree Search (MCTS) methods \cite{mcts}.\footnote{
For a different attempt at understanding the success of UCT, see \cite{selman:uct}.}
It has been argued that this  success is the result of \emph{adaptive sampling methods}: sampling methods that achieve a good 
% HG
% exploration-exploitation tradeoff. Yet, adaptive sampling methods like the well-known Real-Time Dynamic Programming  (RTDP) algorithm  
% have been used before in planning \cite{barto:rtdp}. For us,  an  important reason  for the success of MCTS  methods is that
exploration-exploitation tradeoff. Yet, adaptive sampling methods like  Real-Time Dynamic Programming  (RTDP)  \cite{barto:rtdp}
have been used before in planning. For us,  another important reason  for the success of MCTS  methods is that
they address a slightly different problem; a problem that can be characterized as:

\begin{enumerate}[1.]
\item {anytime} \emph{action selection} over MDPs (and related models) given a time window, resulting in 
good selection when the window is short, and  near to optimal selection when window is sufficiently large, along with 
\item non-exhaustive  search combined with the ability to  use  informed \emph{base policies} for improved performance.
\end{enumerate} From this perspective, an algorithm like RTDP fails on two grounds: first, RTDP does not
appear to make best use of short time windows in large state spaces;  second, and more importantly, 
RTDP  can use admissible heuristics but not informed base policies.  On the other hand, algorithms 
like Policy Iteration \cite{howard:pi}, deliver all of these features except one: they are exhaustive, and thus even to get started, 
they need vectors  with the size of the state space. At the same time, while there are non-exhaustive versions 
of (asynchronous) Value Iteration such as RTDP, there are no similar `focused' versions of Policy Iteration 
ensuring  anytime optimality. Rollouts and nested rollouts are two practical and focused versions of Policy  Iteration 
over large spaces \cite{bertsekas:rollouts,diaconis:nested}, yet neither one aims at optimality.\footnote{
Nested rollouts can deliver optimal action selection  but for impracticable large levels of nesting.}

In this work, we  introduce a new, simple heuristic search algorithm designed to  address points  1 and 2 above, 
and compare it with UCT. We call the new algorithm \emph{Anytime AO*}, because it is a
very simple variation of the classical AO* algorithm for  AND/OR graphs \cite{nilsson:book}
that is optimal even 
in the presence of non-admissible heuristics. Anytime AO* is related  to recent 
anytime and on-line heuristic search algorithms for OR graphs \cite{likhachev:ara,hansen:anytime,koenig:2009,thayer-ruml:anytime}.
It is well known that A* and other best-first algorithms can be  easily converted into anytime optimal algorithms
in the presence of non-admissible heuristics by simply continuing  the search after the first solution is found. The same trick, however,
does not work for  best-first algorithms over AND/OR graphs that must be able to expand leaf nodes of 
the explicit graph that are not part of the best partial solution.
Thus, Anytime AO* differs from AO* in two main points: first, with probability $p$,
it expands leaf nodes that are not part of the best partial solution; second, 
the search finishes when time is up or there are no more leaf nodes to expand at all
(not just in the best solution graph). Anytime AO*  delivers an optimal policy eventually 
and can also use \emph{random heuristics that are not admissible and can be sampled}, 
as when the heuristics result from rollouts of a given \emph{base policy}.  
% We also test Anytime AO* for action selection over large  \emph{infinite-horizon} MDPs that cannot be solved with existing heuristic search
% and dynamic programming algorithms, and  compare it with UCT, both in the presence of informed and non-informed (random) base policies.

\Omit{
The paper is organized as follows. We review MDPs, AO* and UCT, and then introduce Anytime AO* and
compare it with UCT. }

\section{MDPs}

Markov Decision Processes are fully observable, stochastic state models.
In the discounted reward formulation, an MDP is given by a set $S$ of states,  
sets $A(s)$  of actions applicable in each state $s$, transition probabilities 
$P_a(s'|s)$ of moving from $s$ into $s'$ when the action $a \in A(s)$ is applied,
real rewards  $r(a,s)$ for  doing action $a$ in the state $s$, and a discount 
factor $\gamma$, $0 < \gamma < 1$. A solution  is a policy $\pi$
selecting an action $\pi(s) \in A(s)$ in each state $s \in S$. A policy is 
optimal if it maximizes the expected accumulated discounted reward.
%  that is always
% bounded by $R/1-\gamma$, where $R$ is the maximum possible reward. 

Undiscounted finite horizon MDPs replace the discount factor $\gamma$
by a positive integer horizon $H$. The policies for such MDPs are 
functions  mapping nodes $(s,d)$ into actions $a \in A(s)$, where $s$
is a state and $d$ is the horizon to go, $0 <  d \leq H$. Optimal
policies maximize the total reward that can be accumulated in $H$
steps. Finite horizon MDPs are acyclic as all actions decrement the 
horizon to go by $1$. 

Undiscounted infinite horizon MDPs are like discounted MDPs but
with discount $\gamma=1$. For such MDPs to have well-defined
solutions, it is common to assume that rewards are negative, and 
thus represent positive costs, except in certain goal states
that are cost-free and absorbing. If these goal states
are reachable from all the other  states  with positive 
probability, the set of optimal policies is well defined.
%%% BB **** CHECK: Last sentence is not clear *****
%%% HG Seems clear to me .. but feel free to fix it

The MDPs above are reward-based. AO* is used normally in a cost
setting, where rewards $r(a,s)$ are replaced by costs $c(a,s)$,
and maximization is replaced by minimization. The two points of
view are equivalent, and we will use one or the other 
when most convenient.

We  are interested in the problem of selecting the 
action to do in the current state of a given infinite horizon MDPs, whether
discounted or undiscounted. This will be achieved by a lookahead that
uses a limited time window to run an anytime optimal algorithm over the version 
of the MDP that results from fixing the horizon.
% The quality of the action selected 
% normally increases with more time or a larger  $H$, becoming optimal when the time and 
% the horizon are large enough.

\section{AO*}

A finite horizon MDP defines an \emph{implicit AND/OR graph} that can be solved
by the well-known AO* algorithm \cite{nilsson:book}. The root node of this graph
is the pair $(s_0,H)$ where $s_0$ is the initial state and $H$ is the horizon, 
while the terminal nodes are of the form $(s,d)$ where $s$ is a state and
$d$ is $0$, or $s$ is a terminal state (goal or dead-end). 
The children of a non-terminal node $(s,d)$ are the triplets $(a,s,d)$ where $a$
is an action in $A(s)$, while the children of a node $(a,s,d)$ are the nodes
$(s',d-1)$ where $s'$ is a possible successor state of $a$ in $s$; i.e.,
$P_a(s'|s) > 0$.
Non-terminal nodes of the form $(s,d)$ are OR-nodes where an action needs to be
selected, while nodes %of the form
$(a,s,d)$ are non-terminal AND-nodes. 

The solutions graphs to these AND/OR graphs are defined in the standard way;
they include the root node $(s_0,H)$, and recursively, one child of every
OR-node and all children of every AND-node.
The value of a solution is defined recursively: the leaves have value equal to $0$, the OR-nodes $(s,d)$ have
value equal to the value of the selected child, and the AND-nodes $(a,s,d)$
have value equal to the cost of the action $a$ in $s$ plus the sum of
the values for the children $(s',d-1)$ weighted by their probabilities $P_a(s'|s)$.
The \emph{optimal solution graph} yields a minimum value to the root node (cost setting),
and can be computed as part of the evaluation procedure, marking the best child (min cost child)
for each OR-node.  This procedure is called \emph{backward induction}.
The algorithm AO* computes an optimal solution graph in a more selective and incremental 
manner, using a heuristic function $h$ over the nodes of the graph that is \emph{admissible} or
\emph{optimistic} (does not overestimate in the cost setting). 
% BLAI: removed for clarity
% for explicating a minimal part of the \emph{implicit AND/OR graph} 
% that characterizes the finite horizon MDP.

AO* maintains a graph $G$ that \emph{explicates} part of the implicit AND/OR
graph, the \emph{explicit graph}, and a second graph $G^*$, the \emph{best 
partial solution},  that represents an optimal solution of $G$
under the assumption that the tips $n$ of $G$ are the terminal nodes 
with values given by the heuristic $h(n)$.
% A node in $G$ is said to be
% \emph{explicated} and an optimal solution for $G$ is called a
% \emph{best partial graph or solution}.
Initially, $G$ contains the root node of the implicit graph only, and 
$G^*$ is  $G$. Then, iteratively, a non-terminal leaf node is selected
from the best partial solution $G^*$,  and the children of this node in 
the implicit graph are explicated in $G$. The best partial solution  of $G$ 
is then revised by applying an incremental form of  backward induction, 
setting the values of the leaves in $G$ to their heuristic values.
The procedure finishes when there are no leaf nodes in the best partial graph $G^*$. 
If the heuristic values are optimistic, the best partial solution $G^*$ is an 
optimal solution to the implicit AND/OR graph, which is partially explicated in $G$.
In the best case, $G$  ends up containing no more nodes than those in the
best solution graph; in the worst case, $G$  ends up explicating all the
nodes in the implicit graph. Code for AO* is shown in Fig.~\ref{fig:ao*}. 
The choice of which non-terminal leaf node in the best partial graph to expand is 
important for performance  but is not part of the algorithm and it 
does not affect its optimality.
%properties. %% Add ref. to LDFS-AO* if accepted? Puede ser; lo vemos if accepted :-)

\begin{figure}[t]
\centering
\resizebox{3.3in}{!}{
  \fbox{\begin{minipage}{3.9in}
  AO*: $G$ is explicit graph, initially empty; $h$ is heuristic function. \\[1.5ex]
  %and best solution graph contain initially node $(s,H)$ only,
  %where $s$ is seed state and $H$ is the horizon. Best solution graph then obtained from $G$
  %by following marked actions recursively from root. Discount factor is $\gamma$ and 
  %heuristic is $h$.
  \noindent \textbf{Initialization}
  \begin{enumerate}[1.]
  \item Insert node $(s,H)$ in $G$ where $s$ is the initial state
  \item Initialize $V(s,H):=h(s,H)$
  \item Initialize best partial graph to $G$
  \end{enumerate}
  \vskip .5em
  \noindent \textbf{Loop}
  \begin{enumerate}[1.]
  \item[4.] Select non-terminal tip node $(s,d)$ in best partial graph.
    If there is no such  node, \textbf{Exit}.
  \item[5.] Expand node $(s,d)$: for each $a\in A(s)$, add node $(a,s,d)$
    as child of $(s,d)$, and for each $s'$ with $P_a(s'|s)>0$, add node
    $(s',d-1)$ as child of $(a,s,d)$. Initialize values $V(s',d-1)$ for
    new nodes $(s',d-1)$ as $0$ if $d-1=0$, or $h(s',d-1)$ otherwise.
  \item[6.] Update ancestors AND and OR nodes of $(s,d)$ in $G$, bottom-up as:
    \begin{alignat*}{1}
      Q(a,s,d) &:= c(a,s) + \gamma\textstyle\sum_{s'} P_a(s'|s)V(s',d-1), \\
      V(s,d) &:= \textstyle\min_{a\in A(s)} Q(a,s,d).
    \end{alignat*}
  \item[7.] Mark best action in ancestor OR-nodes $(s,d)$ to any action $a$ such 
    that $V(s,d) = Q(a,s,d)$, maintaining marked action if still best.
  \item[8.] Recompute best partial  graph by following marked actions.
  \end{enumerate}
  \end{minipage}}
}
\caption{AO* for Finite Horizon Cost-based MDPs.} 
\label{fig:ao*}
\end{figure}

%% I think this below can be skipped for now ..
\Omit{
The explicit graph $G$ in AO* is an actual graph rather than a tree, as a
node can have multiple parents.
For this reason, backward induction is implemented using a queue that is
initially filled with the new tip nodes that are added to $G$ after each
expansion. Sometimes, a more efficient implementation of AO* is
obtained by storing only the OR-nodes in the explicit graph while
handling the AND-nodes implicitly.}

\Omit{
AND/OR graph incrementally, it keeps the information of the AND-nodes implicit
in the value updates, and does not make them explicit. **** CHECK: SURE? ****
The situation is different in UCT where AND-nodes $(a,s,d)$ associated with the
MDP need to be maintained explicitly. This is because UCT unlike AO* does not
need a full model of the MDP; it just requires a simulator. 
}

\section{UCT}

UCT has some of the flavour of AO* and it is often presented as a best-first
search algorithm too.\footnote{This is technically wrong though, as a best-first
algorithm  just expands nodes in the best partial solution in correspondence to
what is called `exploitation'. Yet  UCT as the other MCTS algorithms, and 
Anytime AO* below, do `exploration' as well.} Indeed, UCT maintains an explicit partial graph 
that is expanded incrementally like  the graph $G$ in  AO*.  The main differences  are in how 
leaf nodes of this graph are selected and expanded, and how heuristic values for leafs are
obtained and propagated.  The code for UCT, which is more naturally described as  a recursive algorithm, 
is shown in Fig.~\ref{fig:uct}.

%In addition, AND-nodes are handled explicitly.
%The code for UCT is shown in Fig.~\ref{fig:uct}. 
%**** CHECK: AND-nodes in UCT ****

\begin{figure}[t]
\centering
\resizebox{3.3in}{!}{
  \fbox{\begin{minipage}{3.9in} % **** CHECK: size of all minipages should be the same ****
    \noindent
    $\hbox{UCT}(s,d)$: $G$ is explicit graph, initially empty;
                       $\pi$ is  base policy;
                       $C$ is exploration constant.
    \begin{enumerate}[1.]
    \item If $d=0$ or $s$ is terminal, Return $0$ 
    \item If node $(s,d)$ is not in explicit graph $G$, then
      \begin{enumerate}[--]
      \item Add node $(s,d)$ to explicit graph $G$
      \item Initialize $N(s,d):=0$ and $N(a,s,d):=0$ for all $a\in A(s)$
      \item Initialize $Q(a,s,d):=0$ for all $a \in A(s)$
      \item Obtain sampled accumulated discounted reward $r(\pi,s,d)$ \\
        by simulating base policy $\pi$ for $d$ steps starting at $s$
      \item Return $r(\pi,s,d)$
      \end{enumerate}
    \item If node $(s,d)$ is in explicit graph $G$, 
      \begin{enumerate}[--]
% HG
%       \item $Bonus(a,s,d)=C\sqrt{2\log N(s,d)/N(a,s,d)}$ if $N(a,s,d)>0$, 
%         or $\infty$ otherwise, for each $a \in A(s)$ 
       \item Let $Bonus(a)=C\sqrt{2\log N(s,d)/N(a,s,d)}$ if $N(a,s,d)>0$, else $\infty$, 
         for each $a \in A(s)$ 
%%%%% **** CHECK: displayed eq for bonus? 
%%%%  HG: ok; as you wish!
\Omit{
      \item Calculate
      \[ bonus(a,s,d)=\left\{\begin{array}{ll}
             C\sqrt{2\log N(s,d)/N(a,s,d)} & \text{if $N(a,s,d)>0$} \\
             \infty & \text{if $N(a,s,d)=0$}
         \end{array}\right. \]
}

%       \item Select action $a = \argmax_{a\in A(s)} [Q(a,s,d) + Bonus(a,s,d)]$
      \item Select action $a = \argmax_{a\in A(s)} [Q(a,s,d) + Bonus(a)]$
      \item Sample state $s'$ with probability $P_{a}(s'|s)$
      \item Let $nv =r(s,a) + \gamma \hbox{UCT}(s',d-1)$
      \item Increment $N(s,d)$ and $N(a,s,d)$
      \item Set $Q(a,s,d):=Q(a,s,d) + [nv - Q(a,s,d)] / N(a,s,d)$
      \item Return $nv$
      \end{enumerate}
    \end{enumerate}
  \end{minipage}}
}
\caption{UCT for Finite-Horizon Reward-based MDPs.}
% Procedure is called over node $(s,H)$ where 
% $s$ is  current state $s$ and  $H$ the horizon. When time runs out, UCT selects 
% applicable action in $s$ that maximizes $Q(a,s,H)$.
\label{fig:uct}
\end{figure}

UCT consists of a sequence of stochastic simulations, like RTDP, that start
at the root node. However, while the choice of successor states is stochastic,
the choice of the actions %in the simulations
is not greedy on the action
$Q$-values as in RTDP, but greedy on the sum of the action $Q$-values and a
bonus term $C\sqrt{2\log N(s,d)/N(a,s,d)}$ that ensures that all applicable
actions are tried in all states infinitely often at suitable rates. 
Here, $C$ is an exploration constant, and $N(s,d)$ and $N(a,s,d)$ are counters
that track the number of simulations that had passed through
the node $(s,d)$ and the number of times that action $a$ has been selected
at such node.

The counters $N(a,d)$ and $N(a,s,d)$ are maintained for the  nodes in the explicit graph 
that is extended incrementally, starting like in   AO*, with the  single root node $(s,H)$. 
The  simulations start  at the root and terminate at a terminal node or at the first node $(s,d)$ 
that is not in the graph. In between, UCT selects an action $a$ that is greedy on the stored value $Q(a,s,d)$ plus the bonus term, 
samples the  next state $s'$ with probability $P_a(s'|s)$, increments the counters $N(s,d)$ and
$N(a,s,d)$, and generates the node $(s',d-1)$. When a node $(s,d)$ is generated that is not in
the explicit graph, the node is added to the explicit graph,  the registers $N(s,d)$, $N(a,s,d)$, 
and $Q(a,s,d)$ are allocated and initialized to $0$, and a total discounted reward $r(\pi,s,d)$ is sampled,
by simulating a base policy $\pi$ for $d$ steps starting at $s$, and  %finally this sampled reward
propagated upward along the nodes in the simulated path. These values 
are not propagated using full Bellman backups as in AO* or Value Iteration, 
but through Monte-Carlo backups that extend the current average with 
a new sampled value \cite{sutton:book}; see Fig.~\ref{fig:uct} for details.

It can be shown that UCT eventually explicates the whole finite horizon MDP
graph, and that it  converges to the optimal policy  asymptotically.
Unlike AO*, however, UCT does not have a termination condition, and moreover,
UCT does not necessarily add a new node to the graph in every iteration, even
when there are such nodes to explicate.
Indeed, while in the worst case, AO* converges in a number of iterations that
is bounded by the number of nodes in the implicit graph, UCT may require an
exponential number of iterations \cite{coquelin-munos,littman:2010}.
On the other hand, AO* is a \emph{model-based approach} to planning that assumes that
the probabilities and costs are known, while UCT is a \emph{simulation-based approach}
that just requires a simulator that can be reset to the current state. 

The differences between UCT and AO* are in the leafs of the graphs selected for expansion, 
the way they are expanded, the values to which  they are initialized, and how the values 
are propagated.
\Omit{
\begin{itemize}
\Omit{
  \item \emph{Nodes in the explicit graph:} AO* maintains OR-nodes $(s,d)$ and
  AND-nodes $(a,s,d)$, while UCT only maintains OR-nodes that stores counters
  $N(s,d)$ and $N(a,s,d)$, and values $Q(a,s,d)$ for $a\in A(s)$.}
\item \emph{Leaf selection:} AO* selects tip nodes $(s,d)$ for expansion from
  the best solution graph. UCT runs a simulation from the root that is greedy
  on the sum of $Q$-values and the exploration bonuses, and adds to the
  explicit graph the first non-terminal node $(s,d)$ that is reached which is
  not in the explicit graph.
\item \emph{Leaf expansion:} AO* expands nodes $(s,d)$ by adding all AND-nodes
  $(a,s,d)$, for $a\in A(s)$, and its successors OR-nodes $(s',d-1)$ for $s'$
  such that $P_a(s'|s) > 0$. UCT adds at most one non-terminal OR-node $(s,d)$
  to the explicit graph in each trial.
\item \emph{Leaf evaluation:} AO* evaluates leafs $(s,d)$ with the heuristic
  $h$ (explicit or sampled). UCT samples the accumulated reward incurred by a
  base policy and uses it to update the $Q(a',s',d')$ values stored at the
  nodes along the sampled path leading to $(s,d)$.
\item \emph{Propagation of values:} AO* propagates values up to ancestors in
  the graph by means of full Bellman backups. UCT propagates values along the
  simulated paths using Monte Carlo backups.
\end{itemize}}
These dimensions are %actually
the ones used to define  a  family of
Monte Carlo Tree Search (MCTS) methods that includes  UCT as the 
best known member \cite{mcts}. The feature that is common to this family of 
methods is the use of Monte Carlo simulations to evaluate the leafs of the 
graph.  The resulting values are heuristic, although not necessarily optimistic
as required for the optimality of AO*.  One  way to bridge the gap between AO* and MCTS methods 
is by modifying AO* to accommodate \emph{non-admissible heuristics}, 
and moreover, \emph{random heuristics that can be sampled} such as
the cost of a base policy.
\Omit{
Indeed, when the heuristic of a node $(s,d)$ is set to the reward
obtained by running a base policy for $d$ steps, 
the \emph{heuristic} is being set to a random variable, whose
samples are not necessarily admissible (even if  the base policy is optimal).
 \footnote{**** CHECK: Left: 
tree vs.\ graph in various methods.***}
}
%%% HG: tree vs.\ graph .. pending 

\section{Anytime AO*}

\Omit{Anytime AO* is a simple variation of AO* aimed at
anytime optimality even in the presence of random
and non-admissible heuristics. By random heuristic, we mean 
a heuristic that corresponds to a random variable that can be sampled, 
such as the cost of a base policy.}

Anytime AO* involves two small changes from AO*. 
The first, shown in Fig.~\ref{fig:anytime-ao*}, is for handling non-admissible heuristics:
rather than always selecting a non-terminal tip node from the
explicit graph that is IN the \emph{best partial graph}, Anytime
AO* selects with probability $p$ a non-terminal tip node from the 
explicit graph that is OUT of the best partial graph. The probability
$p$ is a given parameter between $0$ and $1$, by default $1/2$.
Of course, if a choice for an IN node is decided (a tip in the best
partial graph)  but there is no such node, then an OUT choice is
forced, and vice versa. Anytime AO* terminates when neither IN or OUT choices
are possible; i.e., when no tip nodes in the explicit graph are left,
or when the time is up. 

It is easy to see that Anytime AO* is optimal, whether the
heuristic is admissible or not, because it terminates when
the implicit %AND/OR
graph has been fully explicated. 
\Omit{
with the
same terminal nodes and values. The Bellman updates ensure that
the rest of the nodes get their correct (optimal) values. 
}
In the worst case, the complexity of Anytime AO* is not worse
than AO*, as AO* expands the complete graph in the worst case
too.

The second change in Anytime AO*, not shown in the figure, is for
dealing with random heuristics $h$.  Basically, when the value $V(s,d)$ of a tip
node $(s,d)$ is set to a heuristic $h(s,d)$ that is a \emph{random variable}, 
such as the reward obtained by following a base policy $d$ steps from $s$, 
Anytime AO* uses \emph{samples} of $h(s,d)$ until the node $(s,d)$ 
is expanded. Until then, a `fetch for value'  $V(s,d)$, which occurs each time that a parent node of $(s,d)$ is 
updated, results in a new sample of $h(s,d)$ which is averaged with the previous ones.
%Thus, the $n$-th `fetch' for $V(s,d)$ when $(s,d)$ is a tip node of the
%explicit graph, returns the sum of all previous $n-1$ samples
%$h_1(s,d), \ldots, h_{n-1}(s,d)$ and the new sample $h_{n}(s,d)$ divided
%by $n$.
This is implemented in standard fashion by incrementally 
updating the value $V(s,d)$ using a counter $N(s,d)$ and the new sample.
These counters are no longer needed when the node $(s,d)$ is expanded, as then
the value of the node is given by the value of its children in the graph.

\Omit{The result is an algorithm that is a tiny variation of AO*, but which at
the same time fits with the general idea underlying Monte Carlo Tree Search
methods of using approximate sampled values to guide an anytime optimal
search for solutions.}

\Omit{
The optimality of Anytime AO* does not depend on the heuristic values of 
tip nodes, which could even change randomly from one iteration to the next.
Like in AO*, however, the better the heuristic values, the better the quality
of Anytime AO*  after a short time window in general. Unlike AO*, however,
better heuristic values do not lead to a smaller explicit graph and
faster termination, as Anytime AO* guarantees optimality only after
the whole implicit graph has been explicated. Since this is also true
of AO* in the worst case, the worst case complexity of both algorithms
is the same.}

\begin{figure}[t]
\centering
\resizebox{3.3in}{!}{
  \fbox{\begin{minipage}{3.9in}
  Anytime AO* with possibly non-admissible heuristic $h$: same code as AO* except
  for extra parameter $p$, $0\leq p\leq 1$, and line~4 in Loop for tip
  selection replaced by  4.1 and 4.2 below.
  \begin{enumerate}[4.1.]
  \item Choose IN with probability $1-p$, else Choose OUT:
    \begin{itemize}
    \item IN: Select non-terminal tip node $(s,d)$ from explicit graph
      $G$ that is IN the best partial graph.
    \item OUT: Select non-terminal tip node $(s,d)$ from explicit graph
      $G$ that is NOT in the best partial  graph.
    \end{itemize}
  \item If there is no node $(s,d)$ satisfying the choice condition,
    make the other choice. If there is no node satisfying either choice
    condition, \textbf{Exit} the loop.
  \end{enumerate}
  \end{minipage}}
}
\caption{Anytime AO*. If the heuristic $h(s,d)$ is random, as 
when representing the cost of a given base policy; see text.}
\label{fig:anytime-ao*}
\end{figure}

\section{Choice of Tip Nodes in Anytime AO*}

AO* and Anytime AO* leave  open the criterion for selecting
the tip node to expand.
\Omit{
%%% : in AO* this node must come from the best
%%% solution graph, in Anytime AO*, it must come from the best solution
%%%%graph with probability $1-p$, and from the other tip nodes with
% probability $p$. In order to evaluate the performance   Anytime AO* 
%% in relation to UCT, we define an heuristic criterion for selecting such nodes.
}
For the experiments, we  use a selection  criterion aimed at selecting the tip
nodes that can have the biggest potential impact in the best partial graph.
For this,    a  function $\Delta(n)$ is introduced that measures the change in the  value of the 
node $n$ that is needed in order to produce a change in the  best partial graph.
%%%
%%% Removed due to space %%%
\Omit{
Nodes that are part 
of the best solution graph have non-negative $\Delta$'s, as their 
$V$ or $Q$ values must increase for then being excluded from the
best solution graph. On the other hand, nodes that are not part
of the best solution graph have non-positive $\Delta$'s, as their
$V$ or $Q$ values must decrease for them becoming part of the 
best solution graph. }
The function $\Delta$ is defined top-down over the explicit graph 
as:

\begin{enumerate}[1.]
\item For the root node,  $\Delta(s,H)=\infty$.
\item For children $n_a = (a,s,d)$ of node $(s,d)$ 
in the best solution graph, $\Delta(n_a)$ is $[V(n)-Q(n_a)]$
if $a$ is \emph{not} the best action in $n$, else is 
$\min_b[\Delta(n),Q(n_b)-V(n)]$, for $b \not= a$. 
\item For children $n_a = (a,s,d)$ of node $(s,d)$ 
that is \emph{not} in the best solution graph, $\Delta(n_a)$ is $\Delta(n) + V(n) - Q(n_a)$.
\item For children $n_{s'} = (s',d-1)$ of node $n_a = (a,s,d)$, $\Delta(n_{s'})$
is $\Delta(n_a)/\gamma P_a(s'|s)$.
\end{enumerate}

%%% Omitted -- we don't want to give this too much importance/prominence
%%%% but rather keep it short and explicit, for completeness only
%%%

The tip nodes  $(s,d)$ that are  chosen for expansion 
are the ones that minimize  the values $|\Delta(s,d)|$.
\Omit{
if the choice is IN, they are selected among the tip nodes
in the best solution graph, if the choice is OUT, among those
not in the best solution graph.  }
Since this  computation is expensive, as  it involves a 
complete traversal of the explicit graph $G$,
we select  $N\geq 1$ tip nodes for expansion at a time. 
This selection  is implemented by using two priority queues during the graph traversal:
one for selecting the best $N$  tips in the solution graph (IN),  and one for selecting the best $N$ tips OUT. 
% HG
The first selected tip  is the one with  min $|\Delta(s,d)|$ in the IN queue with  probability $1-p$ and in the OUT queue otherwise.
Once selected, the tip is removed from the queue, and the process is repeated $N$ times.% to collect the $N$ tips.

\section{Experimental Results}

\begin{table*}[t]
\noindent
\centering
\resizebox{6.7in}{!}{
  \begin{minipage}{7.3in}
    \small
    \begin{tabular}{l@{}rr@{}r rr rrr rrr}
      \multicolumn{1}{c}{}
            &
            & \multicolumn{2}{c}{br.\ factor}
            & \multicolumn{2}{c}{UCT--CTP}
            & \multicolumn{3}{c}{random base policy}
            & \multicolumn{3}{c}{optimistic base policy} \\
      \cmidrule(lr){3-4} \cmidrule(lr){5-6} \cmidrule(lr){7-9} \cmidrule(lr){10-12}
      \multicolumn{1}{l@{}}{prob.}
            & \multicolumn{1}{r}{$P(\text{bad})$}
            & \multicolumn{1}{r@{}}{avg}
            & \multicolumn{1}{r}{max}
            & \multicolumn{1}{c}{UCTB}
            & \multicolumn{1}{c}{UCTO}
            & \multicolumn{1}{c}{direct}
            & \multicolumn{1}{c}{UCT}
            & \multicolumn{1}{c}{AOT}
            & \multicolumn{1}{c}{direct}
            & \multicolumn{1}{c}{UCT}
            & \multicolumn{1}{c}{AOT} \\
%%    \midrule
%%    $N=100$ \\
      \midrule
%%    prob    P(bad)       branching             UCTB             UCTO     direct(rand)        uct(rand)        aot(rand)      direct(opt)         uct(opt)         aot(opt)
      10-1  & $19.9$ & $ 8.5$ & $32$ &    $114.8\pm3$ &$\bf 101.3\pm3$ &    $324.7\pm5$ &    $108.4\pm1$ &    $103.1\pm1$ &    $102.8\pm1$ &    $106.1\pm1$ &\G  $102.2\pm1$ \\
      10-2  & $45.6$ & $ 6.5$ & $64$ &    $102.5\pm2$ &    $ 99.9\pm2$ &    $254.8\pm4$ &    $101.1\pm1$ &    $ 98.3\pm1$ &    $145.9\pm2$ &\G$\bf97.9\pm1$ &    $ 99.1\pm1$ \\
      10-3  & $21.9$ & $ 8.9$ & $16$ &    $127.2\pm3$ &$\bf 115.2\pm3$ &    $313.2\pm4$ &    $127.1\pm1$ &    $125.3\pm1$ &    $125.0\pm1$ &    $118.7\pm1$ &\G  $116.5\pm1$ \\
      10-4  & $ 1.4$ & $11.4$ & $32$ &    $ 58.0\pm2$ &    $ 53.4\pm2$ &    $276.4\pm5$ &    $ 55.9\pm1$ &    $ 54.4\pm1$ &    $ 53.0\pm1$ &    $ 53.6\pm1$ &\G$\bf52.5\pm1$ \\
      10-5  & $22.7$ & $ 7.2$ & $16$ &    $ 89.3\pm2$ &$\bf  86.1\pm2$ &    $224.4\pm3$ &    $ 91.3\pm1$ &    $ 90.0\pm1$ &\G  $ 86.7\pm1$ &    $ 89.3\pm1$ &    $ 94.4\pm1$ \\
      10-6  & $24.9$ & $ 7.1$ & $32$ &    $111.5\pm3$ &$\bf  92.6\pm2$ &    $225.4\pm3$ &    $121.4\pm1$ &    $102.9\pm1$ &    $105.0\pm1$ &    $105.4\pm1$ &\G  $ 96.0\pm1$ \\
      10-7  & $ 4.1$ & $11.0$ & $32$ &    $ 83.3\pm2$ &$\bf  66.7\pm1$ &    $244.8\pm4$ &    $ 83.9\pm1$ &    $ 69.9\pm1$ &    $118.4\pm2$ &    $ 74.5\pm0$ &\G  $ 69.0\pm0$ \\
      10-8  & $14.1$ & $ 8.0$ & $32$ &    $ 87.0\pm1$ &$\bf  74.6\pm1$ &    $230.9\pm3$ &    $107.1\pm1$ &    $ 78.3\pm0$ &\G  $ 75.0\pm0$ &    $ 81.8\pm0$ &    $ 76.1\pm0$ \\
      10-9  & $28.1$ & $12.7$ & $32$ &    $ 72.0\pm3$ &    $ 66.5\pm2$ &    $177.6\pm4$ &    $ 71.5\pm1$ &    $ 66.7\pm1$ &\G$\bf63.7\pm1$ &    $ 68.1\pm1$ &    $ 66.7\pm1$ \\
      10-10 & $31.1$ & $ 9.4$ & $32$ &    $ 76.7\pm2$ &    $ 75.4\pm2$ &    $200.0\pm4$ &    $ 79.5\pm1$ &\G$\bf74.5\pm1$ &    $ 76.9\pm1$ &    $ 75.2\pm1$ &    $ 82.7\pm1$ \\
      total &        &      &        &        $922.3$ &$\bf     831.7$ &       $2472.2$ &        $947.2$ &        $863.4$ &        $952.4$ &        $870.6$ &\G      $855.2$ \\
%%
%% uctb=   c(114.8, 102.5, 127.2,  58.0,  89.3, 111.5,  83.3,  87.0,  72.0,  76.7)
%% ucto=   c(101.3,  99.9, 115.2,  53.4,  86.1,  92.6,  66.7,  74.6,  66.5,  75.4)
%% dir_rnd=c(324.7, 254.8, 313.2, 276.4, 224.4, 225.4, 244.8, 230.9, 177.6, 200.0)
%% uct_rnd=c(108.4, 101.1, 127.1,  55.9,  91.3, 121.4,  83.9, 107.1,  71.5,  79.5)
%% aot_rnd=c(103.1,  98.3, 125.3,  54.4,  90.0, 102.9,  69.9,  78.3,  66.7,  74.5)
%% dir_opt=c(102.8, 145.9, 125.0,  53.0,  86.7, 105.0, 118.4,  75.0,  63.7,  76.9)
%% uct_opt=c(106.1,  97.9, 118.7,  53.6,  89.3, 105.4,  74.5,  81.8,  68.1,  75.2)
%% aot_opt=c(102.2,  99.1, 116.5,  52.5,  94.4,  96.0,  69.0,  76.1,  66.7,  82.7)
%% c(sum(uctb), sum(ucto), sum(dir_rnd), sum(uct_rnd), sum(aot_rnd), sum(dir_opt), sum(uct_opt), sum(aot_opt))
%%
      \midrule
%%    prob    P(bad)       branching             UCTB             UCTO     direct(rand)        uct(rand)        aot(rand)      direct(opt)         uct(opt)         aot(opt)
      20-1  & $17.9$ & $13.5$ & $128$&    $210.7\pm7$ &    $169.0\pm6$ &   $1000.3\pm1$ &    $216.4\pm3$ &    $187.4\pm3$ &    $191.8\pm0$ &    $180.7\pm3$ &\G$\bf163.8\pm2$\\
      20-2  & $ 9.5$ & $15.7$ & $64$ &    $176.4\pm4$ &$\bf 148.9\pm3$ &    $676.6\pm1$ &    $178.5\pm2$ &    $167.4\pm2$ &    $202.7\pm0$ &    $160.8\pm2$ &\G  $156.4\pm1$ \\
      20-3  & $14.3$ & $15.2$ & $128$&    $150.7\pm7$ &$\bf 132.5\pm6$ &    $571.7\pm1$ &    $169.7\pm4$ &    $140.7\pm3$ &    $142.1\pm0$ &    $144.3\pm3$ &\G  $133.8\pm2$ \\
      20-4  & $78.6$ & $11.4$ & $64$ &    $264.8\pm9$ &    $235.2\pm7$ &    $861.4\pm1$ &    $264.1\pm4$ &    $261.0\pm4$ &    $267.9\pm0$ &    $238.3\pm3$ &\G$\bf233.4\pm3$\\
      20-5  & $20.4$ & $15.0$ & $64$ &    $123.2\pm7$ &    $111.3\pm5$ &    $586.2\pm1$ &    $139.8\pm4$ &    $128.3\pm3$ &    $163.1\pm0$ &    $123.9\pm3$ &\G$\bf109.4\pm2$\\
      20-6  & $14.4$ & $13.9$ & $64$ &    $165.4\pm6$ &$\bf 133.1\pm3$ &    $670.7\pm1$ &    $178.0\pm3$ &    $160.0\pm2$ &    $193.5\pm0$ &    $167.8\pm2$ &\G  $135.5\pm1$ \\
      20-7  & $ 8.4$ & $14.3$ & $128$&    $191.6\pm6$ &    $148.2\pm4$ &    $862.9\pm1$ &    $211.8\pm3$ &    $170.2\pm2$ &    $171.3\pm0$ &    $174.1\pm2$ &\G$\bf145.1\pm1$\\
      20-8  & $23.3$ & $15.0$ & $64$ &    $160.1\pm7$ &$\bf 134.5\pm5$ &    $704.0\pm1$ &    $218.5\pm4$ &    $154.6\pm3$ &    $167.9\pm0$ &    $152.3\pm3$ &\G  $135.9\pm2$ \\
      20-9  & $33.0$ & $14.6$ & $128$&    $235.2\pm6$ &    $173.9\pm4$ &    $783.4\pm1$ &    $251.9\pm3$ &    $213.7\pm2$ &    $212.8\pm0$ &    $185.2\pm2$ &\G$\bf173.3\pm1$\\
      20-10 & $12.1$ & $15.3$ & $64$ &    $180.8\pm7$ &    $167.0\pm5$ &    $825.7\pm1$ &    $185.7\pm3$ &    $180.5\pm3$ &    $173.2\pm0$ &    $178.5\pm3$ &\G$\bf166.4\pm2$\\
      total &        &      &        &       $1858.9$ &       $1553.6$ &       $7542.9$ &       $2014.4$ &       $1763.8$ &       $1886.3$ &       $1705.9$ &\G$\bf  1553.0$ \\
      \bottomrule
    \end{tabular}
  \end{minipage}
}
\caption{\small  CTP instances with 10 and 20 nodes. $P(\text{bad})$ is probability (in percentage) of
  the instance not being solvable,
  and max branching factor is $2^m$ where $m+1$ is the number of edges with a common node in CTP graph. 
  UCTB and UCTO are  two domain-dependent UCT implementations \cite{malte:ctp}, $\pi$ refers to the base
  policy in our domain-independent UCT and  AOT algorithms, whose  base performance is shown as well. 
  UCT run for 10,000 iterations and AOT for 1,000. Boldface figures are best in whole table; gray cells show
  best among domain-independent implementations.}
\label{table:results:ctp}
\end{table*}

We have evaluated Anytime AO* (abbreviated AOT)  vs.\ UCT as an action selection mechanism
over a number of MDPs. In each case, we run the planning algorithm
for a number of iterations from the current state $s$, 
apply the best action according to the resulting $Q(a,s,H)$ values,  where $H$ is the planning
horizon, and repeat the loop from the  state that results until the goal is reached. 
The \emph{quality profile} of the algorithms is given by the average cost to the 
goal  as a function of the time window for action selection. 
Each data point in the profiles (and table) is the average over 1,000 sampled episodes
that finish when the goal is reached or after  100 steps. For the profiles, the $x$-axis 
stands for the average time per action selected, and the $y$-axis  for the average cost to the goal.
AOT was run with parameter  $p=0.5$ (i.e., equal probability for IN and OUT choices),
and UCT with the exploration `constant' $C$ set to current $Q(a,s,d)$ value of the node;
a choice that appears to be   standard \cite{fern:uct,malte:ctp}. 
The $N$ parameter for the computation of the $\Delta$'s was 
set as a fraction $k$ of the number of iterations, $k=1/10$.\footnote{The number of tip nodes 
in UCT is bounded by the number of iterations (rollouts). In AOT, the number 
of tips is bounded in the worst case by the number of iterations (expansions) multiplied by 
$|A||S|$,  where $|A|$ and $|S|$ are the number of actions and states.}  
The actual codes for UCT and AOT differ from the ones shown
in that they deal with graphs instead of trees; i.e., there are duplicates
$(s,d)$ with the same state $s$ and depth $d$. The same applies to the computation
of the $\Delta$'s.
All experiments were run on Xeon 'Woodcrest' computers of 2.33 GHz and 8 Gb of RAM. 

\subsubsection{CTP.} The  Canadian Traveller Problem  is  a path
finding problem over a graph whose edges (roads) may be blocked
 \cite{ctp:intro}. Each edge is blocked with a given prior probability,
and the status of an edge (blocked or free) can be sensed noise-free 
from either end of the edge. The problem is an acyclic POMDP that
results in a belief MDP whose states are given by the agent position, 
and edge  beliefs that can take 3 values: prior, known to be blocked, and
known to be free.  If the number of nodes is $n$ and the number of edges
in the graph is $m$, the number of MDP states is  $n \times 3^m$.
The problem has been addressed recently using a 
domain-specific implementation of UCT \cite{malte:ctp}. 
We consider instances with 20 nodes from that paper
and instances  with 10 nodes obtained from the authors.
Following  Eyerich,  Keller and Helmert, the actions are  identified with moves
to any node in the  frontier of the known graph. The horizon $H$ can then 
be set to the number of nodes in the graph.

\begin{figure*}
\centering
%trim=left bot right top
\begin{tabular}{ccc}
  \includegraphics[width=3in,clip=on,trim=.75cm .25cm 1.5cm 1cm]{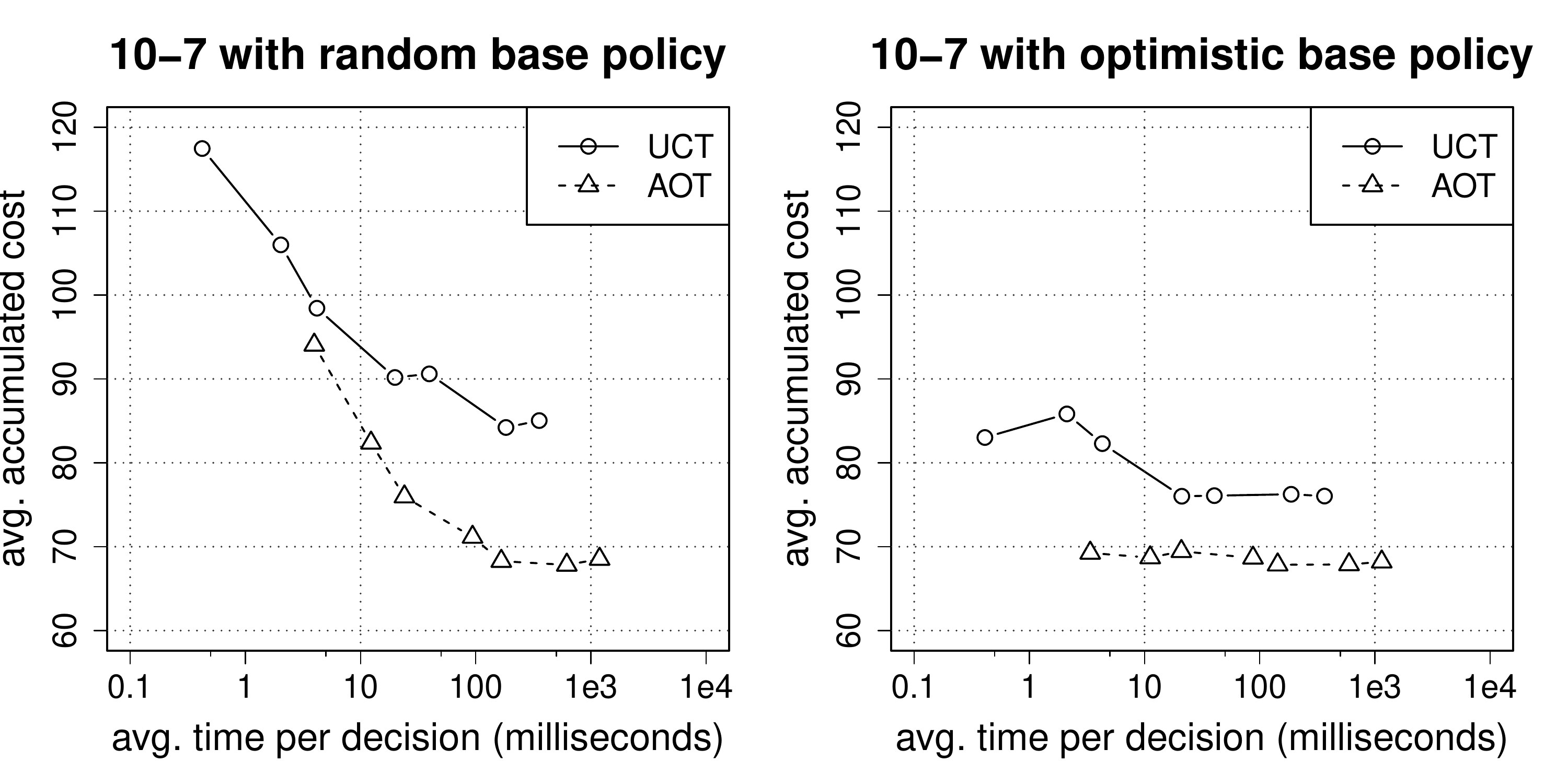} & \quad\quad\quad &
  \includegraphics[width=3in,clip=on,trim=.75cm .25cm 1.5cm 1cm]{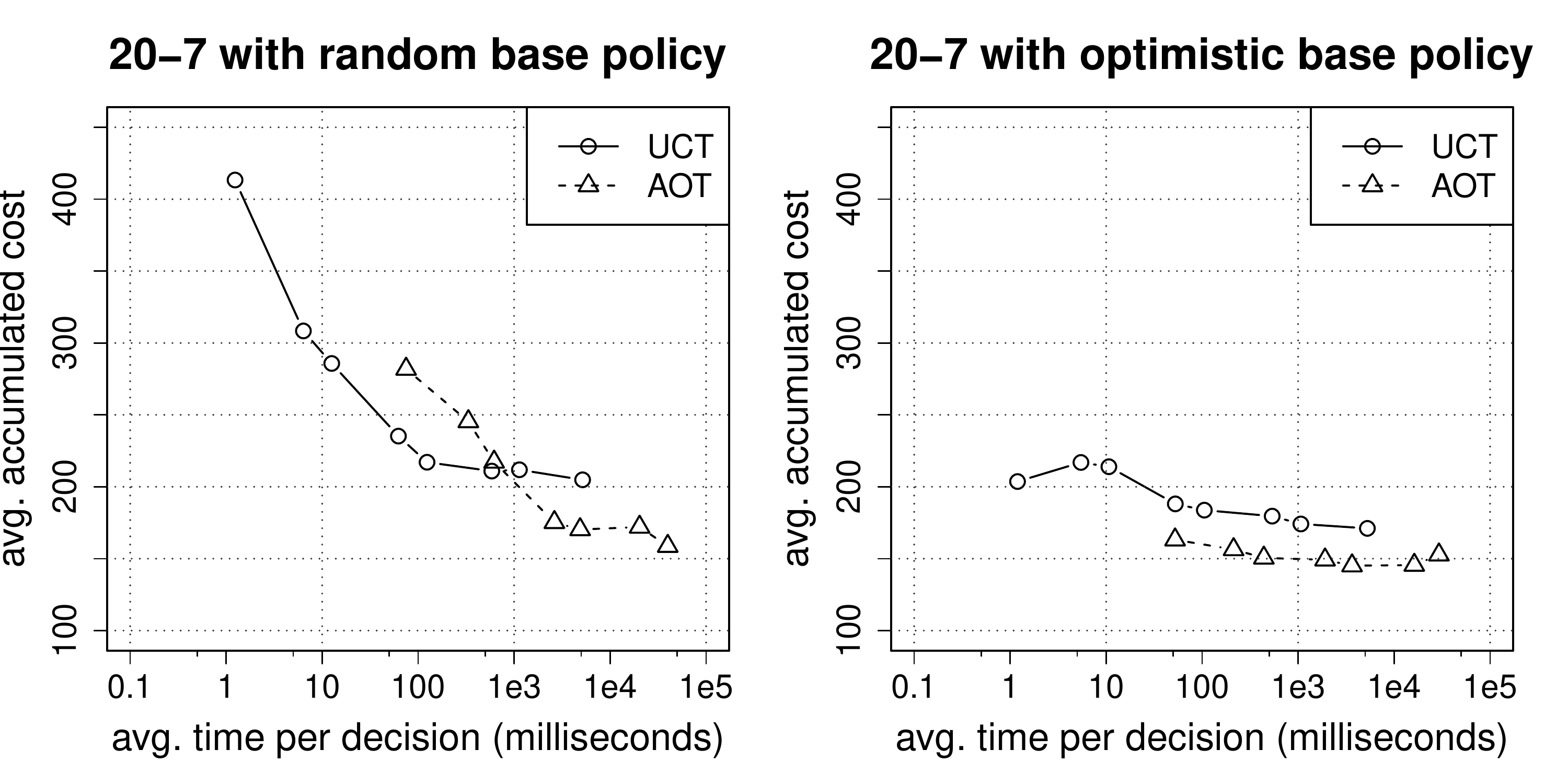} \\
  (a) 10-7 & & (b) 20-7
\end{tabular}
\caption{Quality profiles for CTP instances 10-7 and 20-7.}
\label{fig:ctp:tradeoff:10-7_and_20-7}
\end{figure*}

Quality profiles are shown in Fig.~\ref{fig:ctp:tradeoff:10-7_and_20-7}
for instances 10-7 and 20-7.
On each panel, results on the left are for UCT and AOT using the base
\emph{random policy}, while results on the right are for the base \emph{optimistic policy}
that assumes that  all edges of unknown status are traversable.
% The CTP problem is undiscounted.
The points for  UCT correspond to running 10, 50, 100, 500, 1k, 5k, 10k and 50k iterations (rollouts), 
resulting in the times shown. The points for  AOT, correspond to running  10, 50, 100, 500, 1k, 5k, 10k iterations (expansions). 
% HG
Points that time out are not displayed. % AOT was run with a value of 
% $N=\#expansions/10$ meaning that $M/10$ are expanded in `parallel', when the total number of expansions
% is $M$. 
The curves show  that AOT  performs better than  UCT
except up to the time window of 1 second in  the instance 20-7 for the random base policy.
\Omit{
window where AOT takes over. The first points for AOT appear later in time that the first ones
for UCT because while 10 iterations of UCT introduce at most 10 nodes in the graph,
10 iterations of AOT introduce up to $|A||S|$ nodes, where $A$ and $S$ are the number of 
actions and states. The curves show however  that the more expensive expansions in AOT 
can pay off.} 
%%%
We have computed the curves for the 20 instances and the pattern is similar (but not shown for lack of space).
A comprehensive view of the results  is shown in Table~\ref{table:results:ctp}.
As a reference, we include also  the results for  two specialized implementations of UCT, 
UCTB and UCTO  \cite{malte:ctp}, that take advantage of the specific MDP structure of the CTP, 
use a more informed base policy, and actually solve a slightly simpler version of the 
problem where the given CTP is solvable. While all the algorithms are evaluated over 
solvable instances, AOT and our UCT solve the harder problem of getting to the target 
location if the problem is solvable, and determining otherwise that the problem is unsolvable.
In some instances, the edge  priors are such that the  probability of an instance not being solvable
is high. This is shown as the probability of `bad weather' in the second column of the table.
In spite of these differences, AOT improves upon the domain-specific implementation
of UCT in several  cases, and practically dominates the domain-independent version.
Since UCTO is the state of the art in CTP, the  results show that 
our domain-independent implementations of both AOT and UCT 
are good enough, and that AOT in particular appears to be
competitive with the state of the art in this domain.

\begin{figure}
\centering
%trim=left bot right top
%\includegraphics[width=3.2in,clip=on,trim=.75cm 1cm 1.5cm 1cm]{plots/sailing.pdf}
\includegraphics[width=3in,clip=on,trim=.75cm 1cm 1.5cm 1cm]{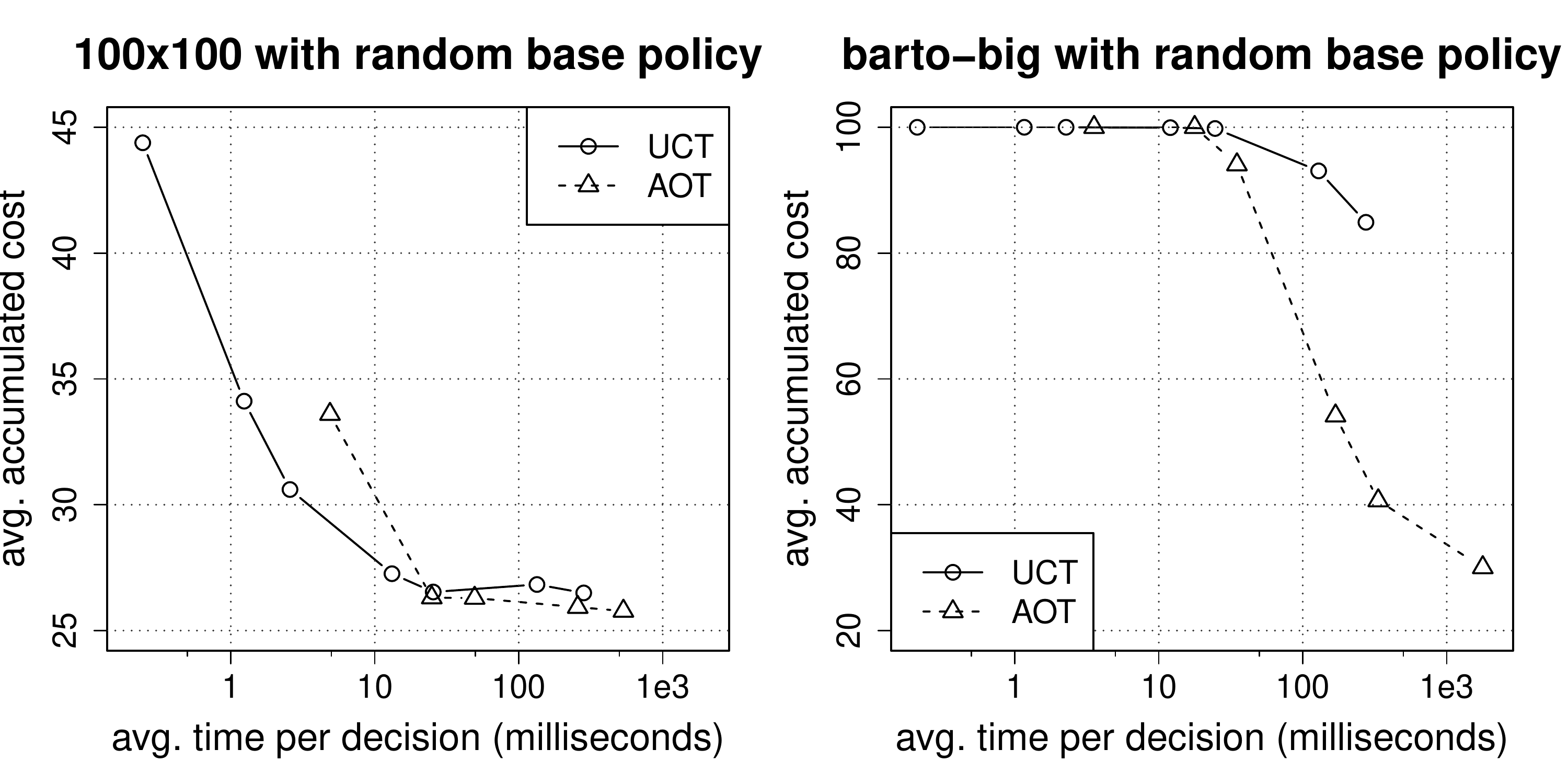}
\caption{Quality profiles for Sailing and Racetrack.}
\label{fig:combined}
\end{figure}

\subsubsection{Sailing and Racetrack.} The Sailing domain is  similar to the one %presented
in the original  UCT paper \cite{uct}. The profile for a $100\times 100$ instance with 
80,000 states  is shown in the left panel of Fig.~\ref{fig:combined} for a random base
policy.
The problem has a discount $\gamma=0.95$ and the optimal value is $26.08$.
UCT and AOT are run with horizon $H=50$.
Profiles for other instances show the same pattern: AOT is slower
to get started because of the more expensive expansions, but then learns
faster.
The profile for the instance 'barto-big' of the Racetrack domain
\cite{barto:rtdp} is shown in the right panel of Fig.~\ref{fig:combined}.
In this case, AOT converges %to the optimal value, $26.13$,
much faster than UCT.
%%%  $N=\#expansions/2$. 
%%The quality profiles are shown in 
% again for the random base policy and the optimistic policy
% that is greedy in the min-min heuristic \cite{bonet:lrtd}.
%%% BLAI
%%%We have computed the curves for several other Sailing instances,
%%%and instances of the  Racetrack domain \cite{barto:rtdp}.
%%%They all follow the same pattern: AOT is slower to get
%%%started because of the more expensive expansions, but
%%%then learns faster. At the same time, the optimistic policies
%%%produce good results in AOT with very few iterations (that take more time
%%%than a few rollouts in UCT), and UCT takes time to improve
%%%the base policies and catch up.

\Omit{
.  The results are shown  in Fig.~\ref{fig:race:tradeoff}
for one of the standard instances. UCT was run for 10, 50, 100, 500, 1k, 5k, 10k iterations,
while AOT for 10, 50, 100, 500, 1k iterations. 
The problem is undiscounted and $H$ is $50$. The same 
base policies have been used. The problem has optimal value
of $26.13430$ and the  number of states is 22,534 states.}

\Omit{
and found that while the $\Delta$
It 
based on the computation of the $\Delta$'s, and
the use of base policies vs.\ heuristic values for
setting the values of tips in AOT. 
In principle, $\Delta$-based selection 
of tip nodes payoffs a lot in comparison with random tip
selection, when a base random policy is used, and less when informed
base policies are used instead  (Fig.~\ref{fig:ctp:selection:10-7}).
}

\subsubsection{Base Policies,  Heuristics, and RTDP.}
\begin{figure}
\centering
%trim=left bot right top
\begin{tabular}{c}
\includegraphics[width=3in,clip=on,trim=.75cm 0cm 1.5cm 1cm]{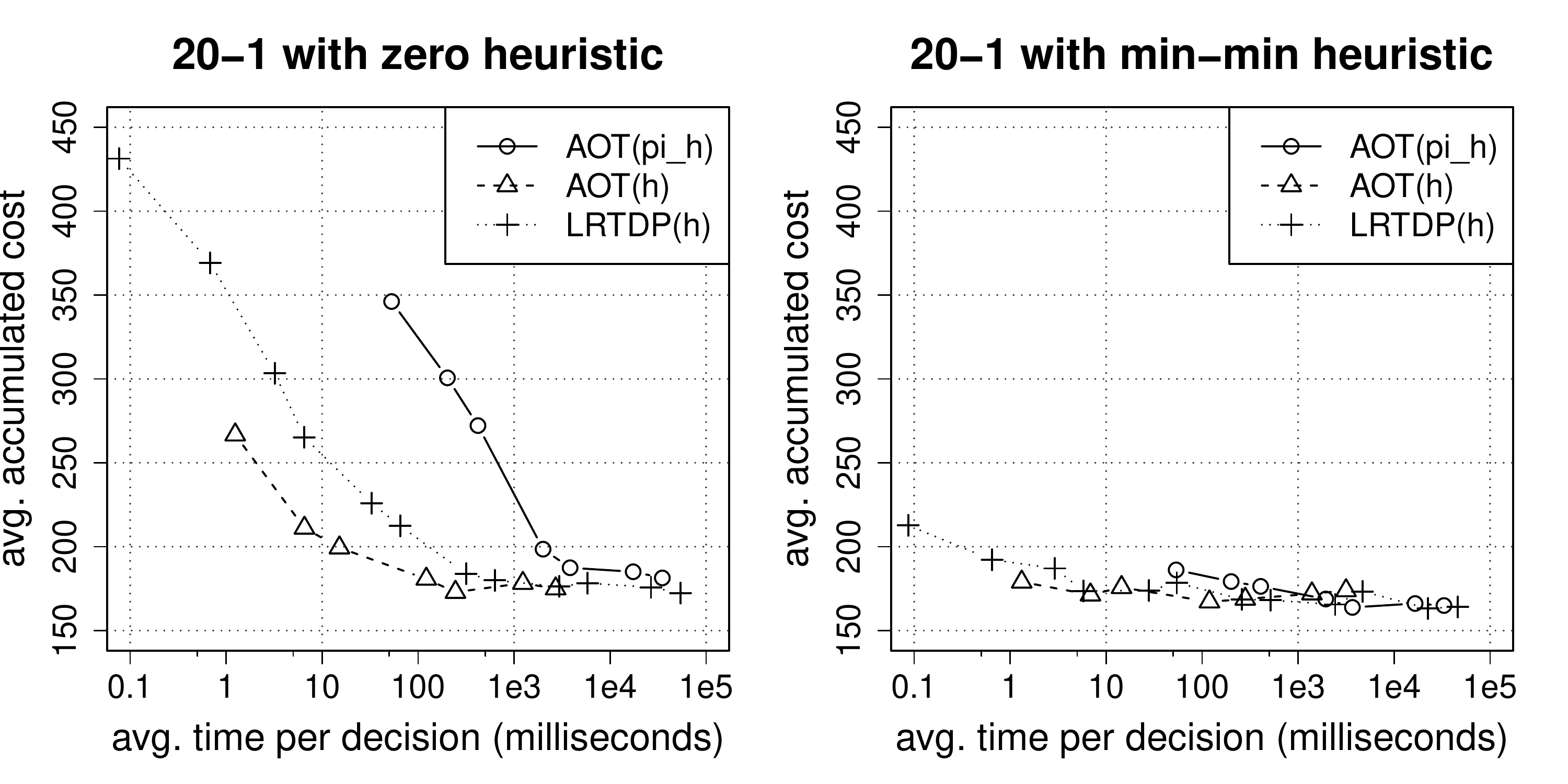} \\
\includegraphics[width=3in,clip=on,trim=.75cm .7cm 1.5cm 1cm]{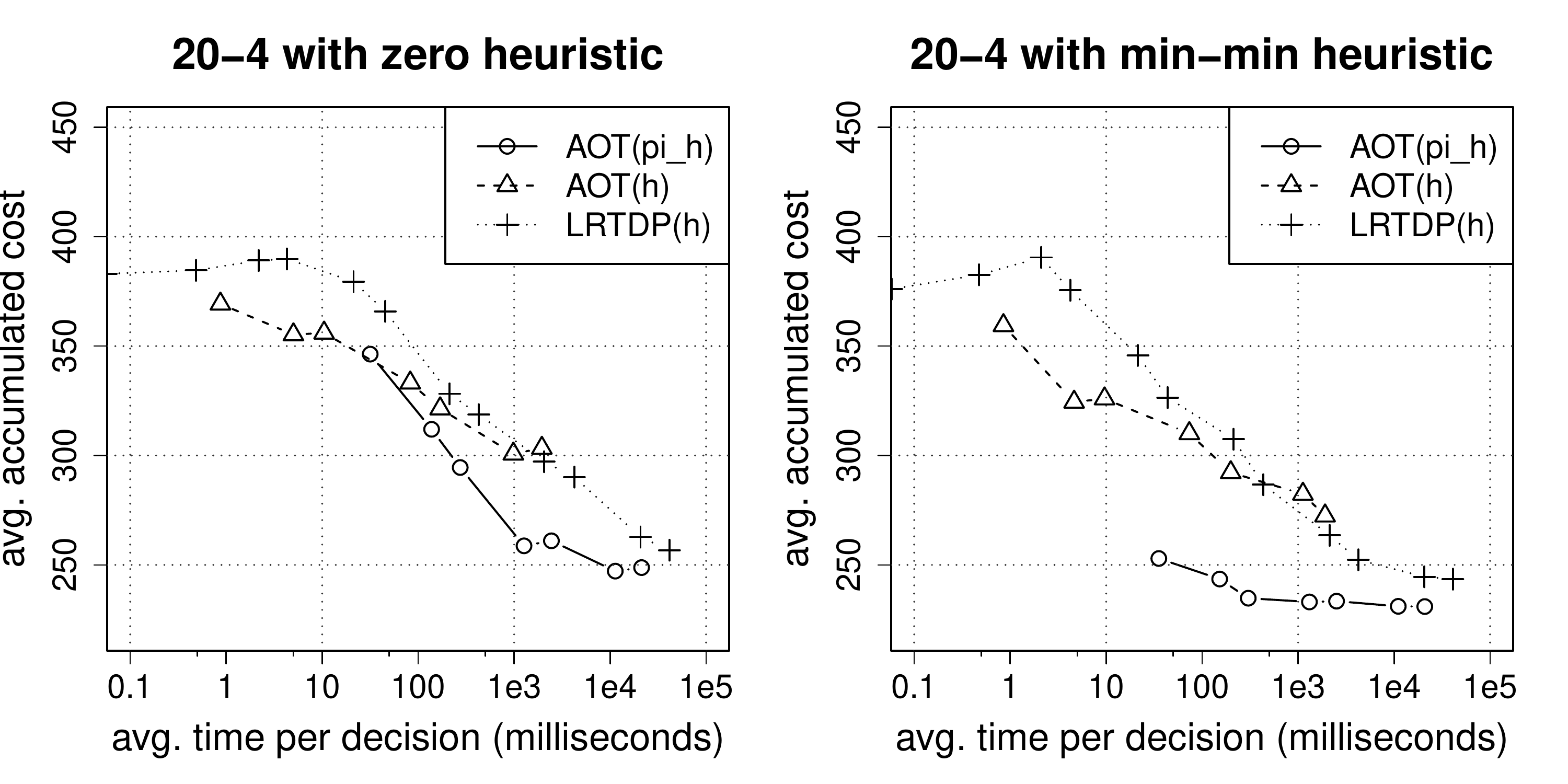}
\end{tabular}
\caption{AOT($\pi_h$), AOT($h$), and LRTDP($h$) on two CTP instances
  for $h=0$ and $h_{min}$. Policy $\pi_h$ is greedy in $h$.}
\label{fig:aot_vs_lrtdp:ctp}
\end{figure}

We also performed experiments comparing the use of heuristics $h$
vs.\ base policies $\pi$ for initializing the value of tip nodes in 
AOT. We  refer to AOT using $\pi$ as a base policy as AOT($\pi$),
and to  AOT using the heuristic $h$ as AOT($h$).
%% NOT SURE: because it will select from OUT is not tip available from IN
%% better say nothing?
%%Notice that AOT($h$) with the parameter $p$ set to $0$, as opposed to $1/2$
%%as in  the experiments, is the standard AO* algorithm.
Clearly, the overhead per iteration is smaller in AOT($h$) than in AOT($\pi$)
that must do a full rollout of $\pi$ to evaluate a tip node. 
Results of AOT($\pi)$ vs.\ AOT($h$) on the instances 20-1 and 20-4 of CTP are shown
in Fig.~\ref{fig:aot_vs_lrtdp:ctp} with both the zero heuristic and the min-min heuristic
\cite{bonet:lrtdp}. The base policy $\pi$  is the policy $\pi_h$ 
that is greedy with respect to $h$ with $\pi_h$ being the random policy when $h=0$.
The curves also show the performance of LRTDP \cite{bonet:lrtdp},  used as an 
\emph{anytime action selection mechanism} that is run over the same finite-horizon MDP as AOT
and with the same heuristic $h$. Interestingly, when LRTDP is used in this form, 
as opposed to an \emph{off-line} planning algorithm, LRTDP does rather well on CTP too, 
where there is no clear dominance  between AOT($\pi_h$), AOT($h$), and LRTDP($h$). 
The curves for the instance 20-1 are rather typical of the CTP instances.
For the zero heuristic, AOT($h$) does  better  than LRTDP($h$) which does better
 than AOT($\pi_h$). On the other hand, for the min-min heuristic, the ranking 
 is  reversed but with differences in performance being smaller. 
There are  some exceptions to this pattern where AOT($\pi_h$) does
better, and even much better than  both  AOT($h$) and LRTDP($h$) for the two heuristics. 
One such instance, 20-4, is shown in the bottom part of  Fig.~\ref{fig:aot_vs_lrtdp:ctp}.

\begin{figure}
\centering
%trim=left bot right top
\includegraphics[width=3in,clip=on,trim=.75cm 1cm 1.5cm 1cm]{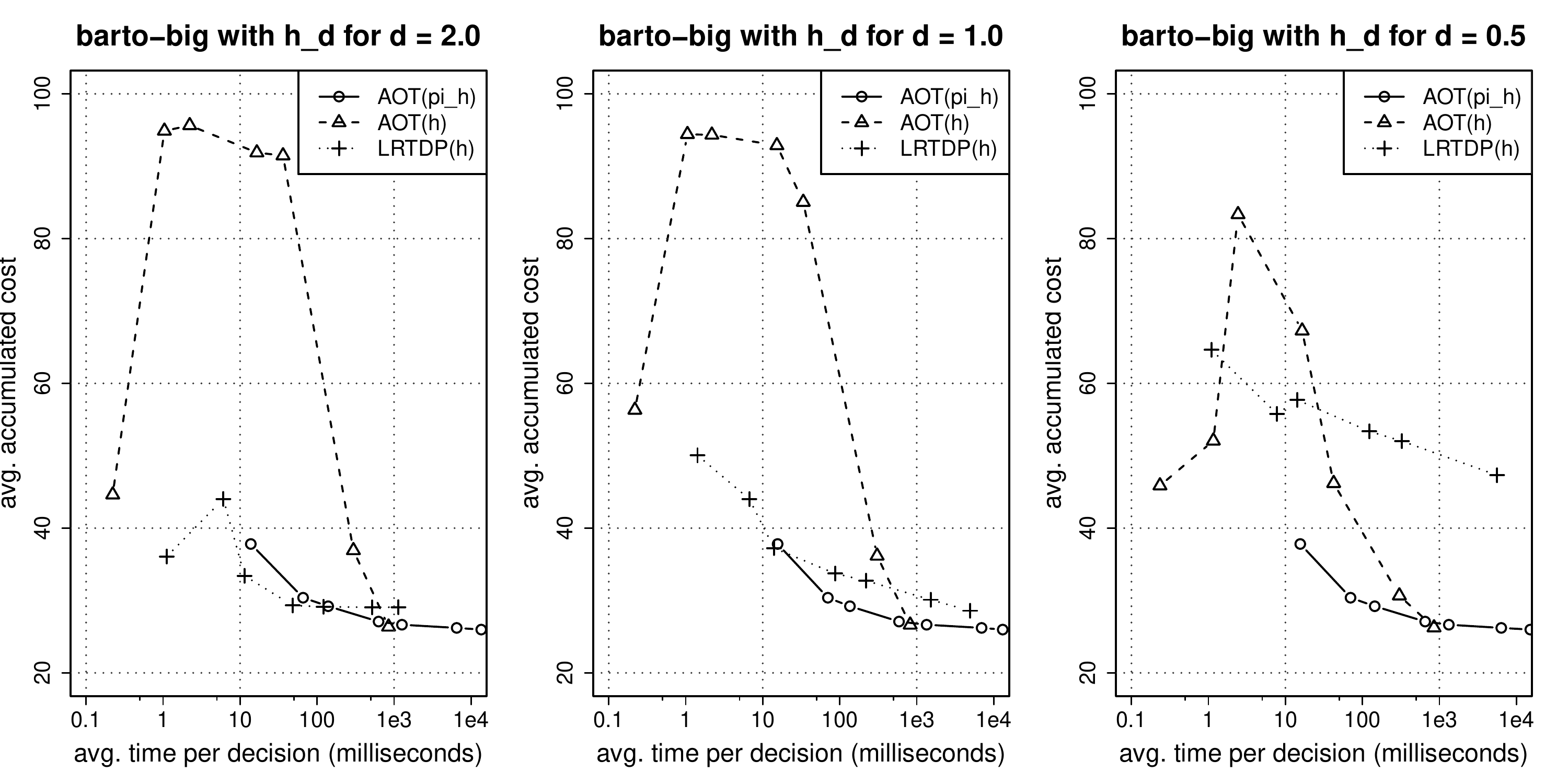}
\caption{AOT($\pi_h$), AOT($h$), and LRTDP($h$) on 
   Racetrack instance for $h=d \times h_{min}$ and $d=2,1,1/2$.}
\label{fig:aot_vs_lrtdp:racetrack}
\end{figure}

Figure~\ref{fig:aot_vs_lrtdp:racetrack} shows the same  comparison for 
a Racetrack instance and three variations $h_d$ of the min-min heuristic
$h_{min}$, $h_d = d \times h_{min}$ for $d=2,1,1/2$. Notice that multiplying
an heuristic $h$ by a constant $d$ has no effect on the policy $\pi_h$
that is greedy with respect to  $h$, and hence no effect on AOT($\pi_{h_d}$).
On the other hand, these changes  affect both AOT($h_d$) and LRTDP($h_d$). 
As expected the performance of LRTDP($h_d$) deteriorates  for  $d < 1$, 
but somewhat surprisingly,  improves for $d > 1$ even as the resulting heuristic
is not (necessarily) admissible.   For small time windows, AOT($h_d$) does worse than LRTDP($h_d$),
but for larger   windows AOT($h_d$) catches up,  and in two cases surpasses  LRTDP($h_d$).
In these cases, AOT with the greedy  base policy $\pi_h$ appears to do  best of all.

\subsubsection{Leaf Selection.} 
\begin{figure}
\centering
%trim=left bot right top
%\includegraphics[width=3.2in,clip=on,trim=.75cm 1cm 1.5cm 1cm]{plots/ctp_selection_10-7.pdf}
\includegraphics[width=3in,clip=on,trim=.75cm 1cm 1.5cm 1cm]{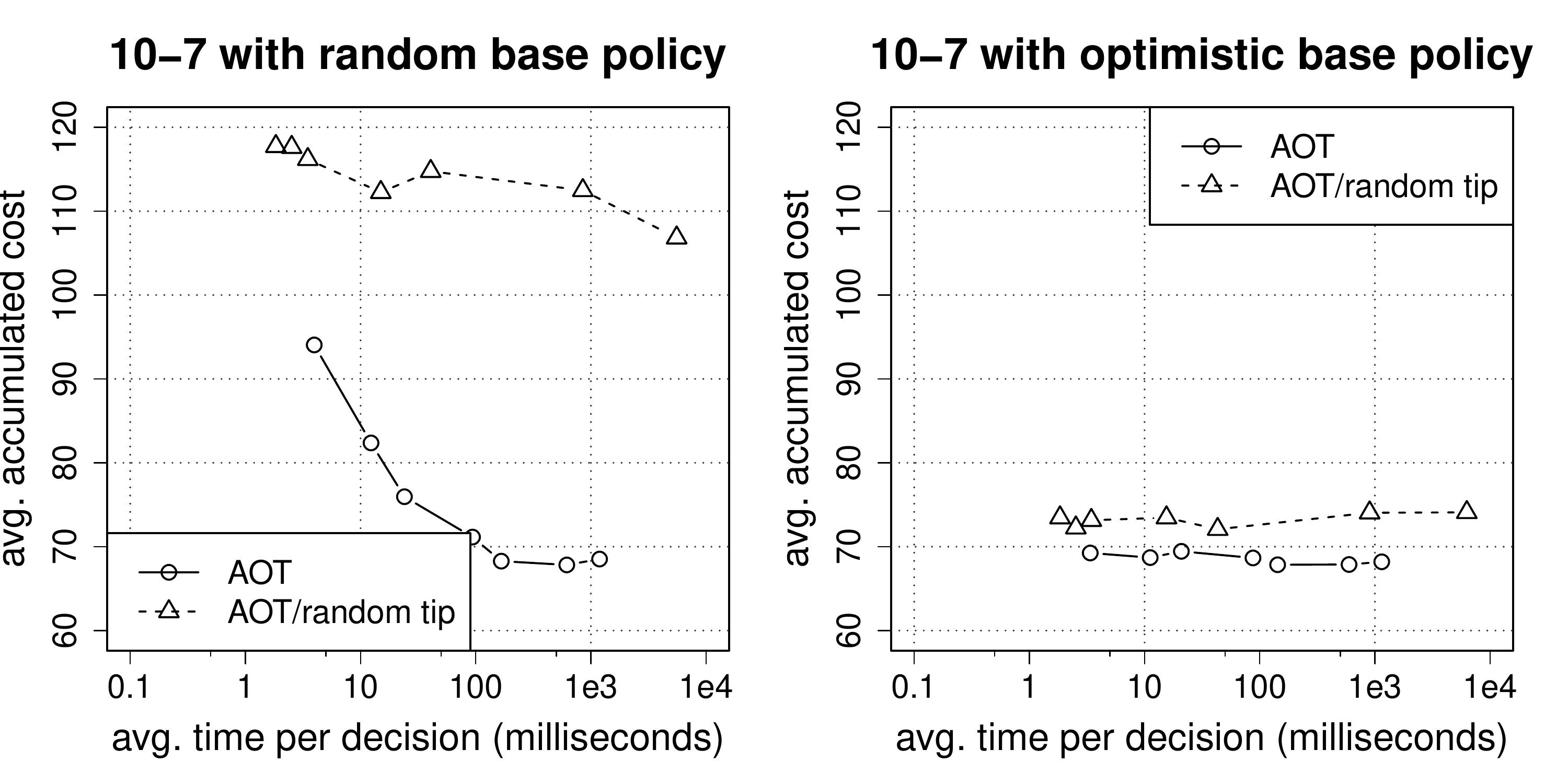}
\caption{Random vs.\ $\Delta$-based tip selection. Problem 10-7.}
\label{fig:ctp:selection:10-7}
\end{figure}

We have also  tested the  value of  the \emph{leaf selection method} used in
AOT by comparing the $\Delta$-based selection of tips vs.\  random tip selection.
It turns out that the former pays off, in particular, when the base policies
are not informed. Some results are shown in Fig.~\ref{fig:ctp:selection:10-7}. 

The results obtained from the experiments above are not conclusive. They show however 
that the new AOT algorithm, while a simple variation of the standard AO* algorithm, appears
to be competitive with both UCT and LRTDP, while producing state-of-the-art results
in the  CTP domain.

\section{Conclusions}

The algorithm UCT addresses the problem of anytime action selection over MDPs and related models,  
combining a  non-exhaustive search with the ability to  use informed base policies. 
In this work, we  have developed a new  algorithm for this task and showed that it compares well with UCT. The new algorithm, Anytime AO* (AOT)
is a very small variation of AO* that retains the optimality of AO* and its worst case  complexity, 
yet it does not require  admissible heuristics and can use base policies. The work
 helps to  bridge the gap between  Monte-Carlo Tree Search (MCTS) methods and \emph{anytime} heuristic search methods,
both of which have flourished in recent years, the former over AND/OR Graphs and Game Trees, the latter
over OR graphs. The relation of Anytime AO* to both classes of methods  suggest also
a number of extensions that are worth exploring in the future. 

\Omit{
 MCTS methods suggest the use of alternative leaf expansion 
strategies for problems with very large branching factors; anytime heuristic search methods
suggest  the use of labels and upper bounds for pruning the search, 
the decomposition of the value function  in the form $g+h$ typical of A* \cite{chakrabarti:g+h,hansen:lao}, 
and the use of a heuristic multiplier parameter $W > 1$ possibly adjustable \cite{richter:w,thayer-ruml:anytime}
for speeding up the search for the goal and subsequent pruning.}

\subsubsection{Acknowledgments.}
H. Geffner is partially supported by grants
TIN2009-10232, MICINN, Spain, and EC-7PM-SpaceBook.

%%%
%%% RTDP
% FSSS Littmann: sampled version of LRTDP for finite horizon and suitable labeling;
%% also too depth-first 

\bibliographystyle{aaai}

\bibliography{control}

\end{document}